\documentclass[letterpaper, 10 pt, conference]{ieeeconf}

\IEEEoverridecommandlockouts

\usepackage{graphicx}
\usepackage{booktabs}
\usepackage{amsmath}
\usepackage{amssymb}
\usepackage{xcolor}
\usepackage{cite}
\usepackage{url}
\usepackage{bm}
\usepackage{xspace}

\usepackage{array}
\usepackage{dblfloatfix}
\usepackage{algorithm}
\usepackage{algpseudocode}

\usepackage{microtype}
\usepackage{hyperref}

\newcommand{\modelname}{CERPE\xspace}

\title{\LARGE \bf
Communication-Efficient Relative Pose Estimation with Vision Foundation Models for Ephemeral Collaborative Perception}

\author{Qihang Li$^{1}$, Jo-Hao Huang$^{1}$, Jiewen Liu$^{1}$, Suyoung Kang$^{2}$, Hao Zhang$^{2}$, and Peng Gao$^{1}$%
\thanks{$^{1}$North Carolina State University, Raleigh, NC 27695, USA. $^{2}$University of Massachusetts Amherst, Amherst, MA 01002, USA. Emails: {\tt\small \{qli39,pgao5\}@ncsu.edu}.}%
}

\begin{document}

\maketitle
\thispagestyle{empty}
\pagestyle{empty}

\begin{abstract}

Relative pose estimation is a fundamental capability for collaborative perception and coordination in multi-robot systems. However, robots encountering each other in real-world environments often operate in short interaction windows and must operate under limited communication bandwidth with intermittent or missing visual overlap caused by occlusions or limited fields of view.
Existing approaches typically rely on global reference frames, assume sustained view overlap, or incur prohibitive communication costs, thereby limiting their applicability to ephemeral collaborative perception.
To address these challenges, we introduce \textit{communication-efficient relative pose estimation} (\textbf{CERPE}), a system-level framework that coordinates vision foundation models to jointly estimate ego-motion and inter-robot relative pose.
CERPE reduces unnecessary raw-observation exchange by using continuously shared fixed-size descriptors to gate event-triggered raw-image requests independently of pose estimation.
Non-overlapping encounters are handled by propagating inter-robot relative poses through metrically scaled ego-motion, thus maintaining relative pose estimates even in the absence of visual overlap.
Experiments in simulation and real-world robots show that CERPE improves 6-DoF relative pose estimation over selected baselines in ephemeral collaborative perception. More details are available on the project website: \href{https://qihangli.com/cerpe/}{https://qihangli.com/cerpe/}.
\end{abstract}

\section{Introduction}
Multi-robot systems have been extensively studied over the past decades due to their ability to enable parallel task execution, expand sensing coverage, and improve robustness through redundancy. They have been deployed across a wide range of real-world applications, including connected autonomous driving \cite{xu2022v2xvit}, multi-robot-assisted warehouse logistics \cite{lifelong_mapf}, and distributed environmental monitoring \cite{gao2023visual}.
To enable effective multi-robot systems, collaborative perception plays a critical role in establishing shared environmental awareness among robot teammates to support downstream tasks such as coordinating actions and improving perception robustness.

\begin{figure}[t]
\vspace{6pt}
    \centering
    \includegraphics[width=\linewidth]{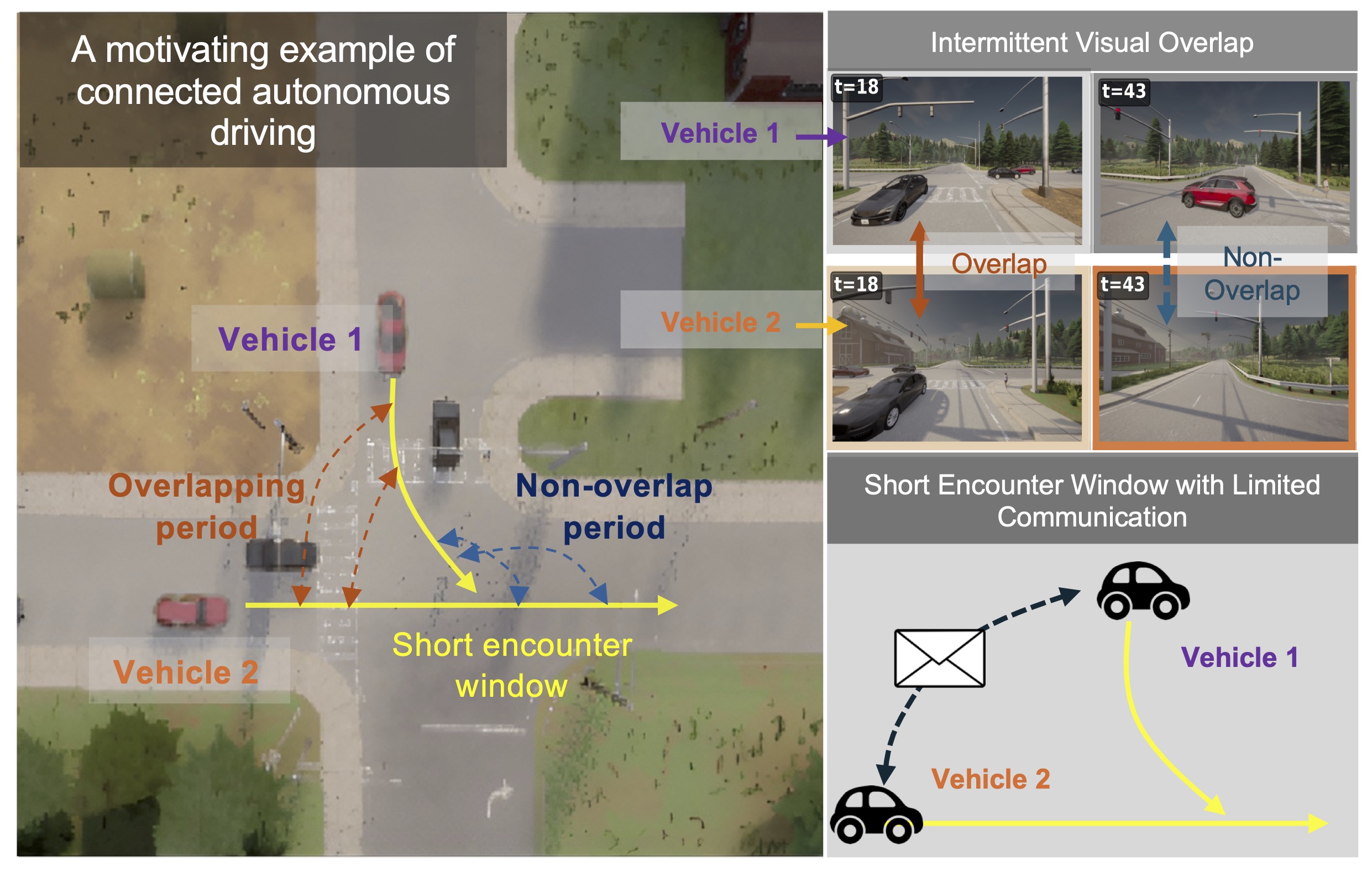}
    \caption{Motivating scenario for communication-efficient relative pose estimation in connected autonomous driving.
   When two connected vehicles encounter each other at an urban intersection, they must estimate their relative pose within a short encounter window to merge observations for collaborative perception, without relying on global references. This requires addressing two key challenges: (1) intermittent visual overlap, which can lead to weak or absent overlapping observations, and (2) limited communication bandwidth, which constrains information exchange during the brief encounter.
    }
    \label{fig:motivation}
\end{figure}

To enable collaborative perception, a fundamental capability is reliable relative pose estimation among robot teammates, which determines the 6-DoF pose of one robot with respect to another using only onboard sensors, without relying on global references such as GPS, motion capture systems, or prebuilt maps.
As shown in Fig.~\ref{fig:motivation}, during an ephemeral encounter at an urban intersection, two connected vehicles must estimate their relative pose within a short interaction window before fusing observations for collaborative perception.
Within this brief encounter, the connected vehicles operate under limited communication bandwidth and may experience weak or even no visual overlap due to occlusions and limited fields of view. These factors make relative pose estimation particularly challenging in real-world scenarios.

Given the importance of relative pose estimation, several approaches have been developed.
Global reference-based methods rely on GPS, indoor tracking systems, or maps, but these references are often unreliable or unavailable \cite{gyagenda2022review}. Collaborative SLAM (CSLAM) methods recover full robot trajectories through optimization over loop closures \cite{cieslewski2018data, tian2022kimera, cao2024multi}, but they are computationally expensive and require maintaining a global map.
Vision-based registration methods estimate relative pose through feature matching or geometric alignment \cite{detone2018superpoint, sarlin2020superglue, gao2021regularized, zhan2024imatching, covis2024, zhang2024diffglue}, yet performance drops under large viewpoint changes.
Recently, vision foundation models have demonstrated strong generalization in feed-forward 3D reconstruction, enabling camera pose estimation by implicitly inferring scene geometry and view transformations from visual observations \cite{wang2024dust3r, wang2025vggt, pi3, mapany, leroy2024grounding}. In multi-robot systems, however, using such representations for inter-robot place recognition or pose estimation would require transmitting raw observations or high-dimensional visual features. This communication cost, together with the common assumption of overlapping observations, limits their effectiveness for short-duration or non-overlapping encounters.

To address these challenges, we propose communication-efficient relative pose estimation (CERPE), which
organizes vision foundation models into a training-free pipeline for metrically scaled ego-motion and inter-robot relative pose estimation under communication constraints during ephemeral collaborative perception. Each robot uses SALAD \cite{salad} to encode observations into fixed-size descriptors and shares them with its local pose in lightweight messages. Descriptor similarity serves as an online proxy for sufficient visual overlap.
For ego-motion estimation, each robot continuously runs VGGT \cite{wang2025vggt} over the current observation and a bounded keyframe memory selected by temporal spacing and descriptor-based viewpoint change. Gate activation triggers raw-observation exchange and cross-robot VGGT inference for direct relative pose estimation. We resolve VGGT's monocular scale ambiguity by aligning its predicted depth with metric depth from Metric3Dv2 \cite{hu2024metric3dv2}. Otherwise, the latest relative pose is propagated using metrically scaled ego-motion from both robots.

The primary contribution of this paper is CERPE, a conditional execution pipeline that gates raw-observation exchange by descriptor similarity, uses metric depth to calibrate monocular scale, and switches to SE(3) ego-motion propagation when visual overlap disappears.
Specifically,
\begin{itemize}

\item CERPE leverages vision foundation models to enable two-stage information sharing with fixed-size descriptor messages for metrically scaled ego-motion and inter-robot relative pose estimation, while explicitly addressing intermittent and missing visual overlap in multi-robot collaborative perception.

\item CERPE supports online deployment for metric-scale 6-DoF relative pose estimation in small multi-robot teams, without relying on global references such as GPS or maps. In our evaluated settings, CERPE reduces position error by up to \textbf{72\%} and rotation error by up to \textbf{90\%} relative to selected baselines across both overlapping and non-overlapping observations.

\end{itemize}

\section{Related Work}
\label{sec:related_work}
\subsection{Collaborative Perception}
Collaborative perception has received increasing attention in recent years due to its ability to overcome the limitations of single-robot perception, particularly in scenarios involving occlusions, long-range sensing, and restricted fields of view~\cite{gao2024survey}.
Collaborative perception has been widely studied for tasks, such as object localization~\cite{liu2020who2com}, detection~\cite{xu2022v2xvit}, trajectory forecasting~\cite{v2vnet}, and connected vehicle control~\cite{cui2022coopernaut} in accident-prone scenarios. By associating observations from multiple viewpoints, these methods build a richer and more robust representation of the environment, leading to improved perception accuracy and reliability.
However, most existing works implicitly assume that robots share sufficiently overlapping observations and can rely on accurate global references (e.g., GPS) to align multi-robot data. In the real-world multi-robot ephemeral encounter scenarios, visual overlap is often intermittent or weak, and global references may be unreliable or even unavailable.

\subsection{Indirect Relative Pose Estimation}
Indirect relative pose estimation refers to methods that first construct explicit intermediate representations such as feature correspondences or geometric alignment, and then recover the relative pose via geometric consistency constraints.

CSLAM-based methods estimate relative poses by maintaining a shared global map and recovering relative transformations through geometric consistency and loop closure constraints~\cite{miller2021any,gao2023visual,gao2020long}. Although effective in structured environments, these approaches rely on persistent map consistency and reliable loop closures, which are difficult to ensure in large-scale or short-duration encounters, thereby limiting their applicability in ephemeral collaborative perception scenarios.
Vision-based indirect methods register observations using feature correspondences~\cite{sun2021loftr,sarlin2020superglue,zhang2024diffglue} or geometric alignment~\cite{jiang2023se,qin2022geometric}. Recent advances incorporate diffusion-based refinement to improve correspondence quality~\cite{wang2023posediffusion,nam2024diffmatch}. Despite improved accuracy, these methods typically assume sufficient visual overlap and degrade under large viewpoint changes or intermittent observations.

\subsection{Direct Relative Pose Estimation}
Direct relative pose estimation refers to methods that directly predict relative poses from visual observations without constructing explicit intermediate representations.
End-to-end regression approaches learn a direct mapping from image pairs to relative poses~\cite{kendall2015posenet,zhang2022relpose,lin2024relpose++,covis2024}. However, they are constrained by their training distributions and typically require overlapping observations, limiting generalization to novel environments or non-overlapping scenarios.

More recently, feed-forward vision foundation models jointly predict scene geometry and camera poses in a single forward pass with strong generalization~\cite{wang2024dust3r,leroy2024grounding,wang2025vggt}. In particular, VGGT~\cite{wang2025vggt} accepts an arbitrary number of input frames, making it a suitable backbone for unifying visual odometry and collaborative perception within a single framework. However, these models predict geometry only up to an unknown scale, rely on high-dimensional features that are expensive to communicate, and assume overlapping observations. These limitations are particularly critical in ephemeral collaborative perception where interaction windows are short and communication bandwidth is constrained.
Our method does not introduce a new foundation model; instead, it contributes a task-specific orchestration protocol over existing models, coupling descriptor-based overlap gating, selective memory insertion, direct inter-robot pose estimation, and SE(3) propagation for intermittent observations.

\begin{figure*}[t]
    \centering
    \includegraphics[width=\textwidth]{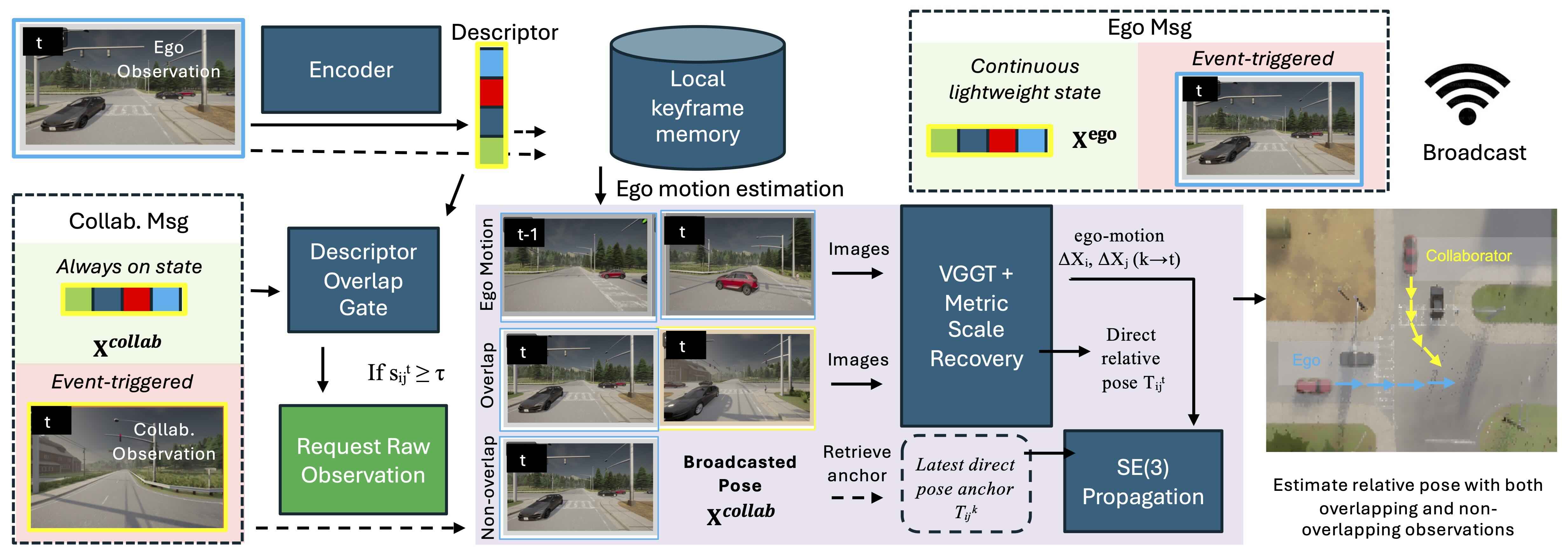}
    \caption{Overview of our \modelname. Each robot continuously broadcasts its local pose and a fixed-size descriptor. Descriptor similarity gates raw-observation requests: gate activation triggers direct inter-robot pose estimation; otherwise, CERPE propagates the latest direct pose anchor using both robots' metrically scaled ego-motion.}
    \label{fig:pipeline}
\end{figure*}

\section{Approach}

\subsection{Problem Formulation}
\label{sec:problem_formulation}
Consider a team of $N$ robots operating in a shared environment. Without loss of generality, we represent a pair of robots as $r_i$ and $r_j$, where $i, j \in \{1, \dots, N\}$ denote robot indices.
Each robot $r_i$ captures an RGB image sequence $\mathcal{I} = \{\bm{I}_i^1, \bm{I}_i^2, \dots\}$ along its trajectory, where $\bm{I}_i^t$ denotes the image captured by robot $r_i$ at time $t$.
Given a pair of image sequences $\mathcal{I}_{i}$ and $\mathcal{I}_{j}$ acquired by robot $r_i$ and $r_j$,
we aim to estimate the relative pose $\bm{T}_{ij}^{t} \in \mathrm{SE}(3)$ between robots $r_i$ and $r_j$ at every time step $t$.
We decompose the relative pose estimation $\phi$ into two sub-problems, including
\begin{itemize}
    \item \emph{Ego-motion estimation.} Each robot maintains a bounded keyframe memory $\mathcal{M}_i^t$ and estimates its local pose $\bm{X}_i^t$ by processing the current observation together with this memory. Since all poses predicted in a VGGT forward pass share the same local coordinate frame, the robot recovers incremental ego-motion by composing the corresponding local poses, e.g., $\bm{T}_i^{t,t-1} = (\bm{X}_i^{t})^{-1}\bm{X}_i^{t-1}$.
    \item \emph{Relative pose estimation.} When robot $r_i$ and robot $r_j$ have visual overlap at time $t$, the relative pose $\bm{T}_{ij}^t$ can be directly estimated from their overlapping observations $(\bm{I}_i^t, \bm{I}_j^t)$ based on vision foundation models~\cite{wang2025vggt}. In non-overlapping periods, the relative pose is propagated using a previously established relative pose estimate and each robot's accumulated ego-motion, which is defined as $\bm{T}_{ij}^{t}=(\bm{X}_j^{t})^{-1}\bm{X}_j^{t-k}
\; \bm{T}_{ij}^{t-k} \; (\bm{X}_i^{t-k})^{-1}\bm{X}_i^{t}$, where $k$ denotes the most recent time step when overlapping observations are available.
\end{itemize}

We address two key challenges of this formulation: (1) enabling vision foundation models with metric scale awareness for ego-motion and relative pose estimation, and (2) overcoming the communication constraints imposed by high-dimensional visual features, thereby enabling relative pose estimation under both overlapping and non-overlapping observations.

\subsection{Metrically Scaled Ego-Motion Estimation}
\label{sec:ego_motion}
To estimate the ego-motion $\bm{T}^{t,t-1}_i = (\bm{X}_i^{t})^{-1}\bm{X}_i^{t-1}$, we adopt VGGT~\cite{wang2025vggt} as the pose estimation backbone, denoted $f_{\mathrm{vggt}}$. At each time step $t$, robot $r_i$ runs the backbone on the current image together with its local memory queue $\mathcal{M}_i^{t-1}$:
\begin{equation}
    \left\{\bm{X}_i^k,\, \bm{D}_i^k\right\}_{\bm{I}_i^k \in \mathcal{M}_i^{t-1} \cup \{\bm{I}_i^t\}} = f_{\mathrm{vggt}}(\mathcal{M}_i^{t-1} \cup \{\bm{I}_i^t\}).
\end{equation}
The predicted poses are expressed in a common local coordinate frame induced by the current memory. Therefore, when the previous frame or keyframe is present in the memory, the ego-motion is recovered by composing the two local poses as $\bm{T}_i^{t,t-1} = (\bm{X}_i^{t})^{-1}\bm{X}_i^{t-1}$; otherwise the robot composes motions through the most recent stored keyframe.
However, as the vision foundation model infers geometry from monocular observations without access to metric depth, the predicted rotations are scale-invariant, whereas the translation component of
$\bm{X}_i^t$ is defined only up to an unknown scale due to the inherent scale ambiguity of monocular vision.
The memory is updated using both temporal spacing and descriptor-based viewpoint change. Let $c_i^t$ denote the cosine similarity between the current descriptor and the descriptor of the most recent local keyframe, and let $\Delta_i^t$ denote their index gap. The current observation is selected as a new keyframe when
\begin{equation}
    (c_i^t < \tau_{\mathrm{m}} \land \Delta_i^t > \Delta_{\mathrm{k}}) \lor c_i^t < \tau_{\mathrm{k}},
\end{equation}
where $\tau_{\mathrm{m}}$ selects frames with moderate viewpoint change after sufficient temporal spacing, and the stricter threshold $\tau_{\mathrm{k}} < \tau_{\mathrm{m}}$ immediately accepts frames with large viewpoint change. The selected keyframes are saved into a bounded memory queue $\mathcal{M}_i^t = \left\{ \bm{I}_k \right\}_{k=1}^{K}$, where $K$ denotes the total number of keyframes in memory.

Given the memory queue $\mathcal{M}_i^t$,
we recover the scale factor $s \in \mathbb{R}^{+}$ for the entire memory by comparing them against metric depth references $\bm{D}^{*}_k$, which are obtained from a metric depth foundation model Metric3Dv2~\cite{hu2024metric3dv2}.
Formally, we compute the metric depth reference as
\begin{equation}
    \bm{D}^{*}_k = f_{\mathrm{m3d}}(\bm{I}_k), \quad \bm{I}_k \in \mathcal{M}_i^t
\end{equation}
where $f_{\mathrm{m3d}}$ denotes the metric depth foundation model.
Then we compute the scale factor as
\begin{equation}\label{eq:scale}
    s(\mathcal{M}_i^t) = \frac{1}{K} \sum_{k=1}^{K} \operatorname{median}_{p} \frac{\bm{D}^{*}_k(p)}{{\bm{D}}_k(p)}.
\end{equation}
where ${\bm{D}}_k(p)$ denotes the depth value at pixel $p$ in the $k$-th keyframe.
$\frac{\bm{D}^{*}_k(p)}{{\bm{D}}_k(p)}$ provides a local estimate of the scale at each pixel. By taking the median over all pixels across all keyframes in the memory, we obtain a stable scale estimate $s$.
The scale factor is then applied to recover the metrically scaled ego-motion as
\begin{equation}
    \bm{T}_i^{t,t-1} = \bm{S}_i \cdot (\bm{X}_i^{t})^{-1}\bm{X}_i^{t-1}, \quad \bm{S}_i = \mathrm{diag}(s_i, s_i, s_i, 1)
\end{equation}
where $\bm{S}_i \in \mathrm{Sim}(3)$ scales the pose from the up-to-scale output to metrically consistent ego-motion.

\subsection{Communication-Efficient Relative Pose Estimation}
\label{sec:collaborative}
To estimate the relative pose $\bm{T}_{ij}^{t}$ between robot $r_i$ and robot $r_j$, we use the same VGGT model as backbone and memory queue to store keyframes.
Once the robot $r_i$ receives the observations sent from its collaborator robot $r_j$,
the relative pose between them is computed as
\begin{equation}\label{eq:relative}
    \bm{T}_{ij},\, \bm{D}_i,\, \bm{D}_j = f_{\mathrm{vggt}}(\bm{I}_i, \bm{I}_j).
\end{equation}
where $\bm{T}_{ij}$ denotes the relative pose between robot $r_i$ and $r_j$, and $\bm{D}_i,\, \bm{D}_j$ denote their respective depth estimates.

However, $\bm{I}_i$ and $\bm{I}_j$ may not overlap due to limited fields of view or occlusion, causing direct estimation to fail. We therefore use descriptor similarity as an online proxy for sufficient visual overlap.
Formally,
given an image $\bm{I}_i$, we utilize the same encoder $f^{\mathrm{enc}}_{\mathrm{vggt}}$ used in VGGT to extract a dense feature map $\bm{F}_i = f^{\mathrm{enc}}_{\mathrm{vggt}}(\bm{I}_i)$. This dense feature map encodes rich semantic and geometric information but is high-dimensional, with spatial resolution proportional to the input image size and channel dimensionality on the order of hundreds. As a result, directly sharing $\bm{F}_i$ between robots would incur prohibitive communication cost and is impractical for bandwidth-constrained multi-robot collaboration.

We therefore adopt two-stage communication: robots frequently exchange fixed-size descriptors to gate candidate overlap, while gate activation triggers raw-observation sharing for inter-robot pose estimation.
Specifically, we adopt the visual aggregation head proposed in SALAD \cite{salad}, which compresses the dense visual feature map into a compact global descriptor by optimally aggregating local patch features through Sinkhorn-based optimal transport \cite{cuturi2013sinkhorn}, while discarding non-informative regions via a learned dustbin mechanism. The compressed descriptor is computed as
\begin{equation}\label{eq:descriptor}
    \bm{d}_i = \psi(\bm{F}_i) \in \mathbb{R}^{D},
\end{equation}
where $\psi$ denotes the visual aggregation head. For fixed $D$ and precision, message size is fixed, while the compression ratio depends on image resolution and encoding. The robots compute cosine similarity $score=\frac{\bm{d}_i^\top \bm{d}_j}{\|\bm{d}_i\| \, \|\bm{d}_j\|}$. A pair with $score \ge \tau$ is treated as a candidate overlap and triggers raw-observation exchange; otherwise the gate remains closed.

\begin{algorithm}[t]
\caption{Robot $r_i$ egocentric relative pose estimation.}
\label{alg:pipeline}
\begin{algorithmic}[1]
\Require Observation $\bm{I}_i^t$; messages $\{\bm{m}_j^t\}$; memory $\mathcal{M}_i^{t-1}$
\Ensure Relative poses $\{\bm{T}_{ij}^t\}$; ego-motion $\bm{X}_i^t$
\Statex \hspace{-\algorithmicindent}\textbf{Feature Extraction \& Communication:}
\State $\bm{d}_i^t \gets \psi\!\bigl(f^{\mathrm{enc}}_{\mathrm{vggt}}(\bm{I}_i^t)\bigr)$ \Comment{Fixed-size descriptor}
\State Broadcast $\bm{m}_i^t = (i,\, \bm{X}_i^t,\, \bm{d}_i^t)$
\For{each received $\bm{m}_j^t = (j,\, \bm{X}_j^t,\, \bm{d}_j^t)$}
    \State $s_{ij}^t \gets \frac{\bm{d}_i^{t\top} \bm{d}_j^t}{\|\bm{d}_i^t\|\,\|\bm{d}_j^t\|}$ \Comment{Descriptor similarity}
    \If{$s_{ij}^t \geq \tau$}
        \State Request $\bm{I}_j^t$; enqueue into $\mathcal{M}_i^t$
    \EndIf
\EndFor
\Statex \hspace{-\algorithmicindent}\textbf{Pose Estimation \& Scale Recovery:}
\State $\{{\bm{X}}_k, {\bm{D}}_k\} \gets f_{\mathrm{vggt}}(\mathcal{M}_i^t \cup \{\bm{I}_i^t\})$
\State $s \gets \tfrac{1}{K} \sum_{k} \operatorname{median}_{p}\, \bm{D}^{*}_k(p) / {\bm{D}}_k(p)$
\State ${\bm{X}}_k \gets s \cdot {\bm{X}}_k, \;\forall k$ \Comment{Metrically scaled poses}
\Statex \hspace{-\algorithmicindent}\textbf{Relative Pose Estimation:}
\For{each robot $r_j$}
    \If{$s_{ij}^t \geq \tau$} \Comment{Candidate overlap}
        \State $\bm{T}_{ij}^t \gets (\bm{X}_j)^{-1} \bm{X}_i$
    \Else \Comment{Non-overlapping}
        \State $\bm{T}_{ij}^t \gets (\bm{X}_j^{t})^{-1}\bm{X}_j^{t\text{-}k}\, \bm{T}_{ij}^{t\text{-}k}\, (\bm{X}_i^{t\text{-}k})^{-1}\bm{X}_i^{t}$
    \EndIf
\EndFor
\Statex \hspace{-\algorithmicindent}\textbf{Memory Management:}
\State $c_i^t \gets s(\bm{d}_i^t,\, \bm{d}_i^{t_\text{last}})$; $\Delta_i^t \gets t - t_\text{last}$
\If{$(c_i^t < \tau_{\mathrm{m}} \land \Delta_i^t > \Delta_{\mathrm{k}}) \lor c_i^t < \tau_{\mathrm{k}}$}
    \State Enqueue $\bm{I}_i^t$ into $\mathcal{M}_i^t$ \Comment{New keyframe}
\EndIf
\If{$|\mathcal{M}_i^t| > L$}
    \State Evict oldest entry from $\mathcal{M}_i^t$
\EndIf
\end{algorithmic}
\end{algorithm}

If the descriptor gate identifies a candidate overlap, the ego robot requests the raw observation from its collaborator. Once the collaborator's observation is received, it is inserted into the ego robot’s memory queue, and the inter-robot relative pose is estimated using Eq.~(\ref{eq:relative}).
If the gate is not activated, the ego robot does not request the raw observation, thereby avoiding unnecessary communication and computation.
The relative pose is propagated using a previously established relative pose estimate $\bm{T}_{ij}^{t-k}$ at time $t-k$ and two robots' accumulated ego-motion, which is computed as
\begin{equation}
    \bm{T}_{ij}^{t}=(\bm{X}_j^{t})^{-1}\bm{X}_j^{t-k}
\; \bm{T}_{ij}^{t-k} \; (\bm{X}_i^{t-k})^{-1}\bm{X}_i^{t}
\end{equation}
where $k$ denotes the most recent time step when overlapping observations are available.
To estimate relative pose in both overlapping and non-overlapping conditions, directly sharing the raw image $\bm{I}$ among robots is bandwidth-intensive. Since overlap occurs only intermittently, CERPE triggers this high-bandwidth transmission only when the descriptor-based gate is activated, reducing unnecessary raw-observation exchange.

Algorithm~\ref{alg:pipeline} summarizes the decentralized execution. Each robot broadcasts its fixed-size descriptor and local pose, requests raw observations only upon gate activation, and otherwise propagates the latest inter-robot estimate using metric ego-motion. Bounded memory retains frames after sufficient temporal and viewpoint change, or immediately after a large viewpoint change.

\section{Experiments}
\subsection{Experiment Setup}

We comprehensively evaluate our \modelname in three experimental settings:
\begin{itemize}
    \item A high-fidelity connected autonomous driving (CAD) simulation based on CARLA~\cite{carla}, which generates connected vehicle trials across 4 intersections, comprising 71 encounter scenarios with a total of 3{,}461 image pairs.
    \item Real-world CAD trials built upon OpenMars~\cite{mars}, which captures multi-robot long-range encounters in urban environments. We use the multi-robot subset and select 23 trajectory pairs (6{,}900 image pairs) that contain camera-level overlap, as many sequences in this dataset have weak visual overlap between robots' views.
    \item A physical robot team of three Agilex Limo Pro robots deployed in 13 manually constructed indoor scenarios that cover large perspective changes, non-overlapping views, occluded views, and dynamic objects. We uniformly select 130 keyframes from 13 trajectories, yielding 382 pairwise image pairs with valid ground-truth poses. Ground-truth poses are obtained from a motion capture system.
\end{itemize}

In all experiments,
we adopt VGGT~\cite{wang2025vggt} as the pose estimation backbone with training-free metric scale recovery defined in Eq.~(\ref{eq:scale}) and pretrained compact descriptors from SALAD~\cite{salad} for place recognition defined in Eq.~(\ref{eq:descriptor}).
In simulation, metric scale is recovered via Metric3Dv2~\cite{hu2024metric3dv2}.
In real-world deployments, onboard depth sensors provide metric depth for scale alignment.
The keyframe memory uses $\tau_{\mathrm{m}} = 0.9$ for moderate viewpoint change, $\tau_{\mathrm{k}} = 0.2$ for large viewpoint change, a fixed minimum index gap $\Delta_{\mathrm{k}}$, the descriptor-gate threshold $\tau = 0.2$, and the memory capacity $L = 5$.

For evaluation, we compare \modelname with three previous methods and one baseline, including
\begin{itemize}
    \item \textbf{SuperGlue}~\cite{sarlin2020superglue} that is a modular feature-matching pipeline that detects and matches sparse keypoints across image pairs. We recover the pose via RANSAC~\cite{fischler1981random}. As SuperGlue cannot recover metric scale, we report its translation performance using ground-truth scale.
    \item \textbf{NOPE}~\cite{nope2025}: an object-level graph matching method that estimates relative pose by exploiting spatial relationships among detected objects (e.g., vehicles and pedestrians) in the scene.
    \item \textbf{Co-VisNet}~\cite{covis2024}: an end-to-end model that directly regresses robot relative pose from image pairs.
    \item A baseline \textbf{\modelname w/o EM} that performs relative pose estimation without using ego-motion propagation to address non-overlapping observations.
\end{itemize}

\begin{table*}[!t]
\centering
\caption{Quantitative results of our \modelname  compared with selected baselines based on position, rotation error and success rate.
}
\label{tab:main_results}
\resizebox{\textwidth}{!}{%
\begin{tabular}{lccccccccc}
\toprule
Method
& \multicolumn{3}{c}{CAD Simulation}
& \multicolumn{3}{c}{Real-world CAD}
& \multicolumn{3}{c}{Physical Robot Teams} \\
\cmidrule(lr){2-4} \cmidrule(lr){5-7} \cmidrule(lr){8-10}
& $e_{\mathrm{pos}}$ (m) $\downarrow$ & $e_{\mathrm{rot}}$ (rad) $\downarrow$ & Succ. (\%) $\uparrow$
& $e_{\mathrm{pos}}$ (m) $\downarrow$ & $e_{\mathrm{rot}}$ (rad) $\downarrow$ & Succ. (\%) $\uparrow$
& $e_{\mathrm{pos}}$ (m) $\downarrow$ & $e_{\mathrm{rot}}$ (rad) $\downarrow$ & Succ. (\%) $\uparrow$ \\
\midrule
SuperGlue~\cite{sarlin2020superglue} & 16.00$^{\dagger}$ & 0.895 & 77.5 & 7.62$^{\dagger}$ & 0.169 & 93.7 & 0.12$^{\dagger}$ & 0.165 & 98.4 \\
NOPE~\cite{nope2025} & 9.41 & 0.776 & 33.4 & 39.41 & 1.682 & 45.0 & --- & --- & --- \\
Co-VisNet~\cite{covis2024} & 16.22 & 0.906 & 100 & 29.95 & 0.271 & 100 & 0.88 & 0.257 & 100 \\
\midrule
\modelname w/o EM & \textbf{5.03\,/\,3.51}$^{\dagger}$ & \underline{0.128} & 100 & \underline{10.27\,/\,5.84}$^{\dagger}$ & \underline{0.063} & 100 & \textbf{0.45\,/\,0.09}$^{\dagger}$ & \underline{0.096} & 100 \\
\modelname  & \underline{8.66\,/\,5.06}$^{\dagger}$ & \textbf{0.089} & 100 & \textbf{8.26\,/\,2.94}$^{\dagger}$ & \textbf{0.063} & 100 & \underline{0.54\,/\,0.11}$^{\dagger}$ & \textbf{0.086} & 100 \\
\bottomrule
\end{tabular}%
}
\vspace{0.5em}

{\raggedright\scriptsize
$^{\dagger}$\,Translation with ground-truth scale. For \modelname, both predicted and GT-scale translations are shown (pred.\,/\,GT).
\par}
\end{table*}
To quantitatively evaluate and compare \modelname with other methods,
we use three metrics, including position error, rotation error, and success rate. Given the estimated translation $\hat{\mathbf{t}}$ and ground-truth translation $\mathbf{t}$, and the estimated and ground-truth rotations represented as unit quaternions $\hat{\mathbf{q}}, \mathbf{q} \in \mathbb{S}^3$, the position and rotation errors are defined as:
\begin{align}
    e_{\mathrm{pos}} &= \|\hat{\mathbf{t}} - \mathbf{t}\|_2, \\
    e_{\mathrm{rot}} &= 4 \arcsin\!\left(\frac{1}{2}\min\left(\|\hat{\mathbf{q}} - \mathbf{q}\|,\, \|\hat{\mathbf{q}} + \mathbf{q}\|\right)\right)
\end{align}
where the rotation metric follows the geodesic distance on $\mathbb{S}^3$ via Roma~\cite{roma}. We additionally report the success rate, defined as the percentage of frames for which a method produces a valid pose estimate. Failures arise when a method cannot recover a pose (e.g., insufficient feature matches or inliers) or when the method determines that the input is outside its operating range (e.g., no detectable overlap).

\subsection{Evaluation in CAD Simulation}
We first evaluate our approach in CAD simulation where two connected vehicles encounter each other at an urban intersection. The encounter window lasts approximately $3-5$ seconds and involves large relative viewpoint changes. As the vehicles traverse the intersection, occlusions caused by surrounding traffic frequently lead to transitions from overlapping to non-overlapping visual observations.

\begin{figure}[!t]
\vspace{4pt}
    \centering
    \includegraphics[width=\columnwidth]{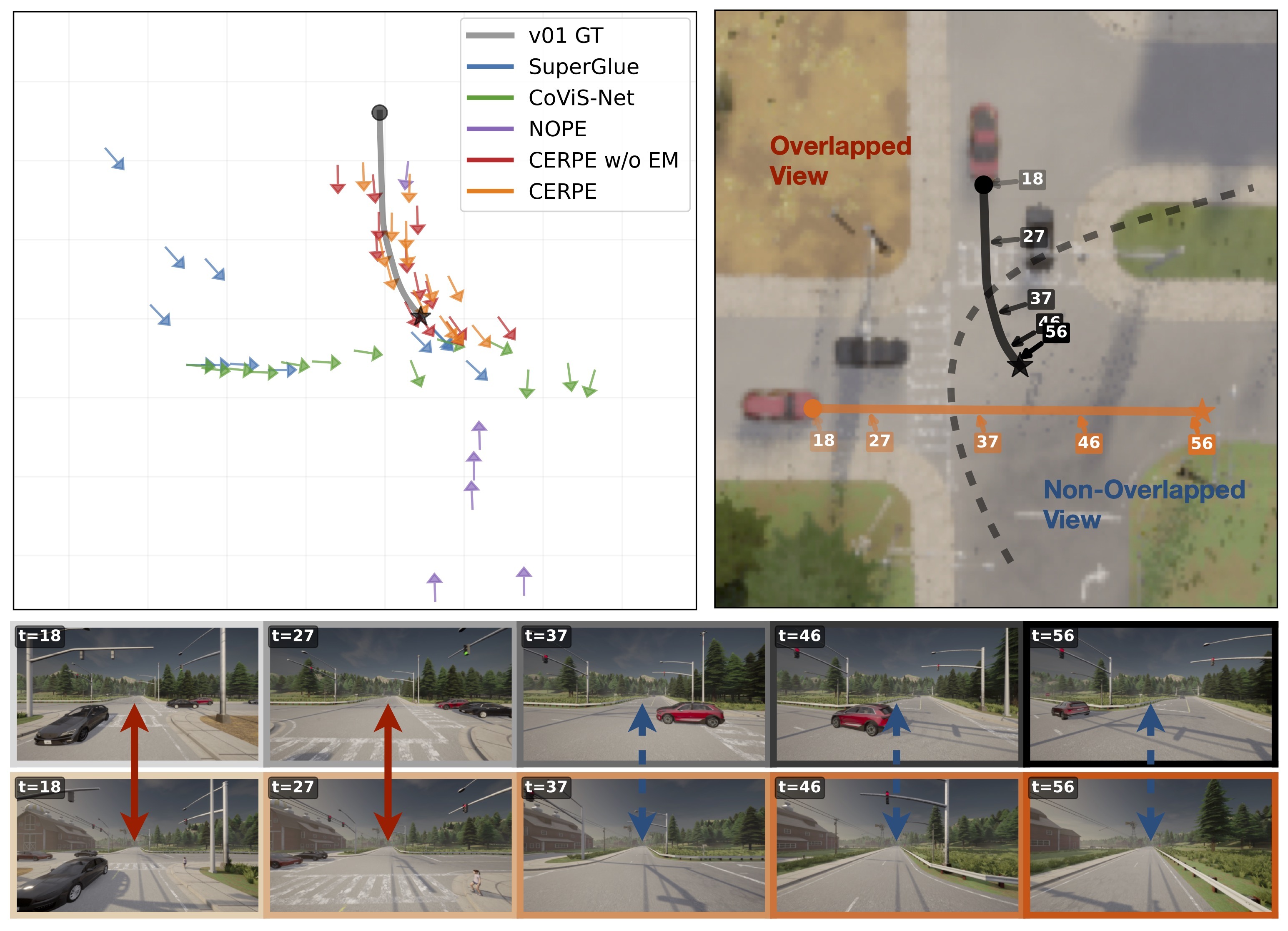}
    \caption{Qualitative results in CAD simulation.
    \textbf{Right:} bird's-eye view of both vehicles' ground truth trajectories, with the dashed line separating overlapping and non-overlapping periods.
    \textbf{Left:} ground-truth and predicted relative poses of the target vehicle (black trajectory) in the reference vehicle's frame (orange trajectory).  \textbf{Bottom:} temporally corresponding image pairs from the two vehicles with red arrows and blue dashed arrows indicating overlapping and non-overlapping views, respectively.}
    \label{fig:quali_carla}
\end{figure}
As shown in Fig.~\ref{fig:quali_carla}, the ego vehicle initially shares overlapping views with its collaborator, but the overlap gradually disappears as the vehicles diverge.
NOPE predicts relative poses in a nearly opposite direction to the ground truth. This is because two visually similar black vehicles appear in both robots' views.
Since NOPE relies on object-level spatial relationships to estimate relative poses, it confuses object identities under such ambiguous visual cues.
Co-VisNet shows a strong bias toward a fixed formation prior, consistently predicting its collaborator at a short distance in front of the ego vehicle, regardless of the true relative pose under large viewpoint changes and partial occlusions.
In contrast, our method maintains temporally stable relative pose estimates throughout the encounter by jointly reasoning over multiple keyframes stored in memory, enabling continued estimation after visual overlap is lost.

The quantitative results are shown in Table~\ref{tab:main_results}. Our method achieves competitive results across the evaluated scenarios while consistently maintaining a 100\% success rate. Among the comparison methods, NOPE achieves a translation error of 9.41\,m but succeeds on only 33.4\% of pairs. Co-VisNet always produces an estimate but with substantially higher errors; and SuperGlue fails on a notable fraction of pairs due to insufficient feature matches.
Our baseline \modelname w/o EM outperforms the comparison methods in CAD simulation, achieving 5.03\,m position error and 0.128\,rad rotation error. Our full approach further improves rotation to 0.089\,rad but increases translation error to 8.66\,m.
This result highlights an operating condition of ego-motion propagation: in high-overlap settings, direct estimates already provide strong translation constraints, while short propagation windows can add ego-motion and scale drift. Thus, propagation is best viewed as a continuity mechanism for missing-overlap periods rather than a module that uniformly improves every metric.

\begin{figure}[!t]
    \centering
    \includegraphics[width=\columnwidth]{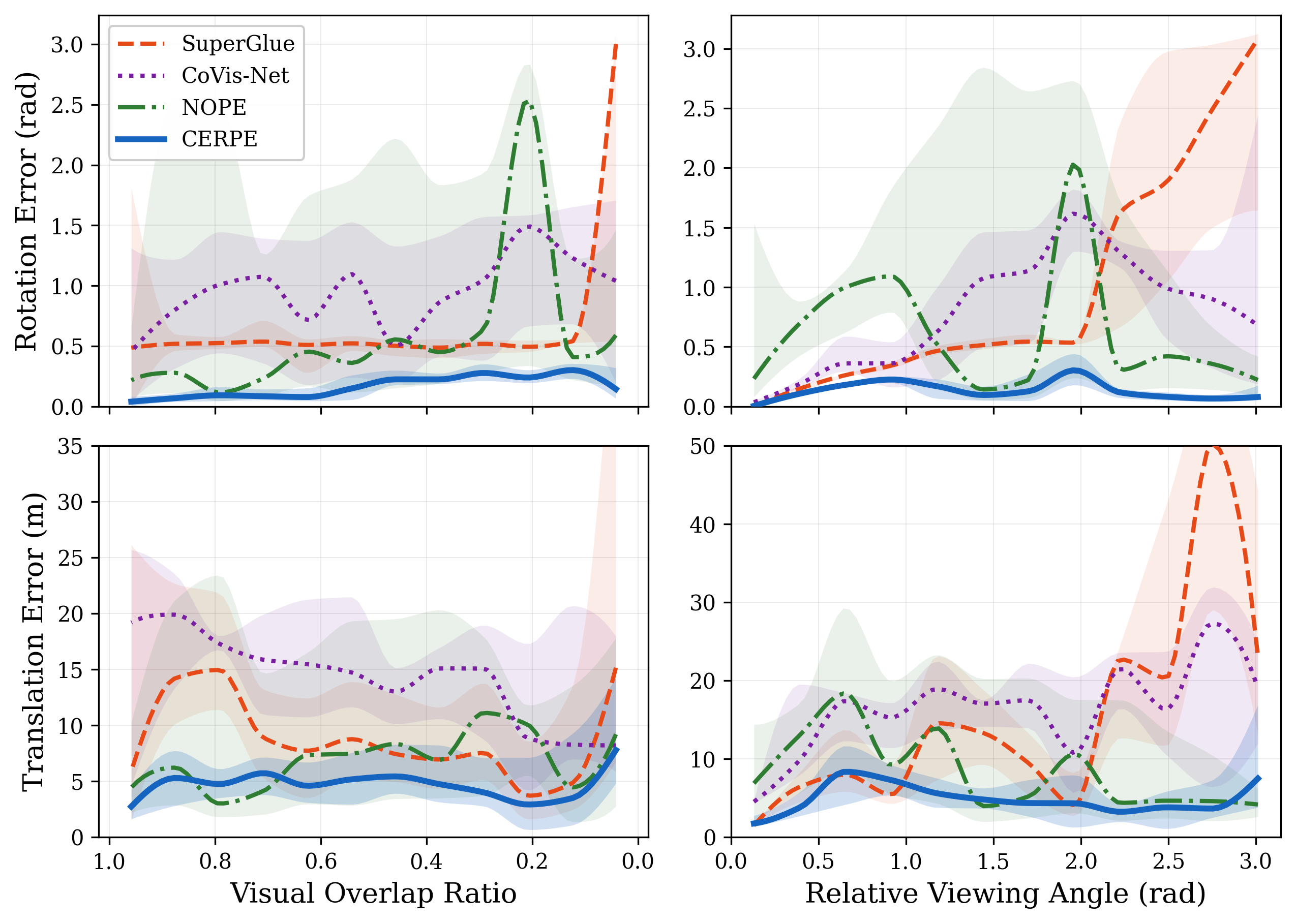}
    \caption{Analysis of performance under varying degrees of visual overlap and viewpoint changes. Compared with selected baselines, our approach remains stable across large variations in visual overlap and view angles.
    }
    \label{fig:error_analysis}
    \vspace{-6pt}
\end{figure}

Fig.~\ref{fig:error_analysis} analyzes how each method's performance varies with visual overlap ratio and relative viewing angle given the CAD simulation. Our method maintains a median rotation error below 0.15\,rad across all overlap bins, whereas SuperGlue's median exceeds 1.5\,rad when overlap falls below 20\%, reflecting its dependence on sufficient feature correspondences. NOPE exhibits an interquartile range spanning over 1\,rad in rotation even at moderate overlap, indicating unstable object-level matching. Co-VisNet achieves comparable rotation accuracy to ours at high overlap but incurs translation errors exceeding 15\,m due to its lack of metric scale. Regarding viewing angle, the comparison methods degrade sharply beyond 2\,rad relative rotation, while our method remains more stable even at viewing angles approaching $\pi$\,rad, indicating improved stability in geometrically challenging regimes.

\subsection{Evaluation in Real-World CAD}

\begin{figure}[!t]
    \centering
    \includegraphics[width=\columnwidth]{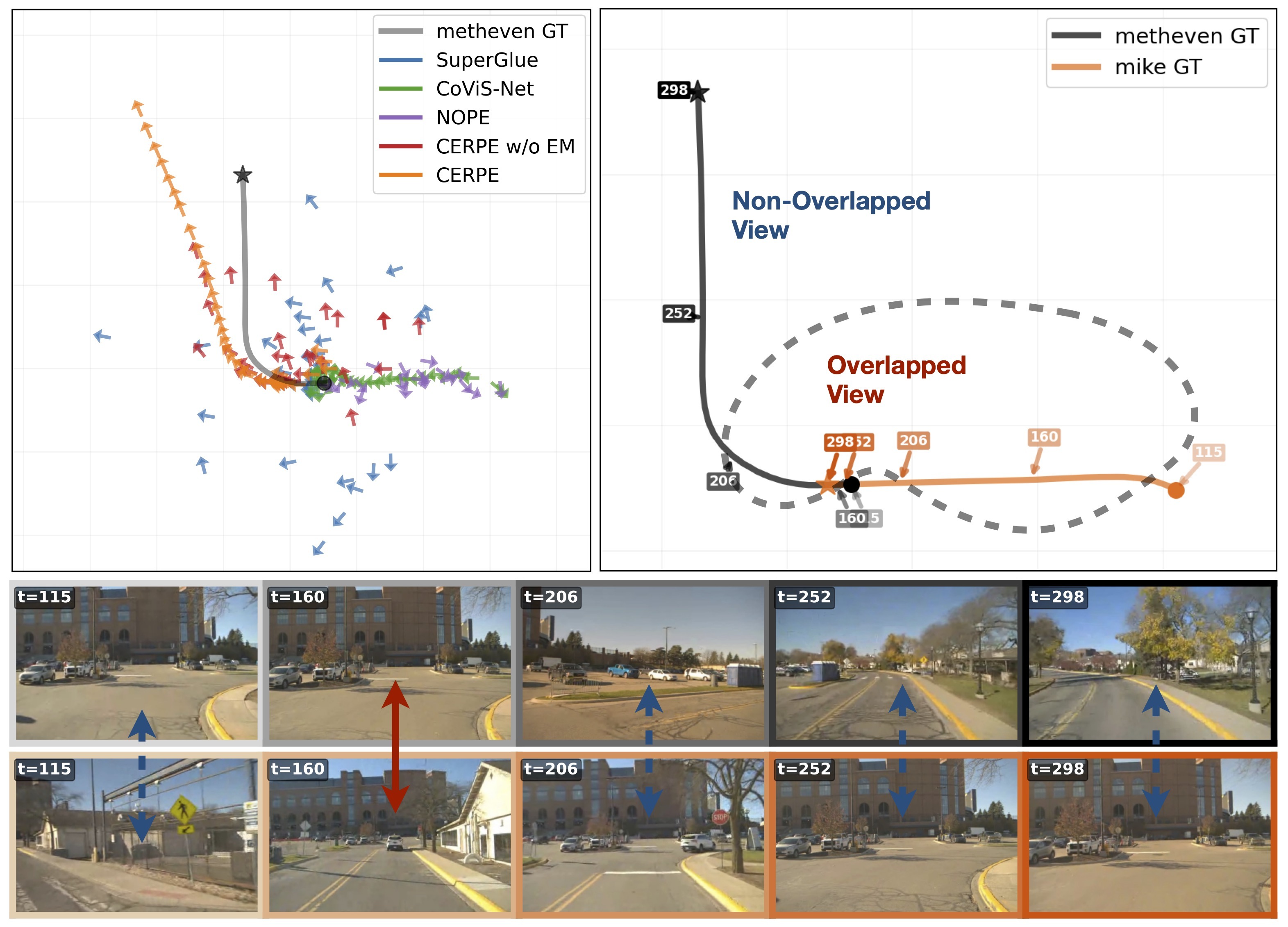}
    \vspace{-6pt}
    \caption{Qualitative results in the real-world CAD. \textbf{Right:} bird's-eye view of both vehicles' trajectories with the gray dashed line separating overlapping and non-overlapping periods. \textbf{Left:} ground-truth and predicted relative poses. \textbf{Bottom:} temporally corresponding image pairs.}
    \label{fig:quali_mars}
\end{figure}

We further evaluate \modelname in the real-world CAD scenario.
Unlike in simulation, where visual overlap between vehicles can be deliberately controlled, real-world driving scenarios typically exhibit much smaller overlap due to unconstrained vehicle motion, long inter-vehicle distances, and occlusions from surrounding traffic.

As shown in Fig.~\ref{fig:quali_mars}, when a low-overlap encounter occurs, our method recovers a lower-error relative pose than the comparison methods by incorporating the inter-robot image into its memory. Moreover, after the overlap ends, our approach maintains temporal continuity through ego-motion propagation combined with the previously established estimate, whereas the other methods either fail entirely or degrade to unconstrained guesses. Notably, although error propagation during non-overlapping periods inevitably introduces some drift, the incorporation of visual odometry improves the temporal consistency of the estimated trajectory, preventing abrupt jumps exhibited by baselines that treat each frame independently. When overlap resumes, the direct estimate corrects accumulated drift.

The quantitative results of real-world CAD are shown in Table~\ref{tab:main_results}.
Compared with simulation, this scenario
features longer trajectories with extended non-overlapping periods. Our \modelname\ achieves lower rotation error of 0.063\,rad than all comparison methods, with the nearest competitor SuperGlue at 0.169\,rad with ground-truth scale. In translation, our \modelname achieves the best predicted-scale result of 8.26\,m, followed by our baseline \modelname w/o EM at 10.27\,m. With ground-truth scale, ours further reduces translation error to 2.94\,m, compared with SuperGlue at 7.62\,m with a 93.7\% success rate.

\subsection{Evaluation on Physical Robot Teams}
\begin{figure}[!t]
\vspace{6pt}
    \centering
    \includegraphics[width=\columnwidth]{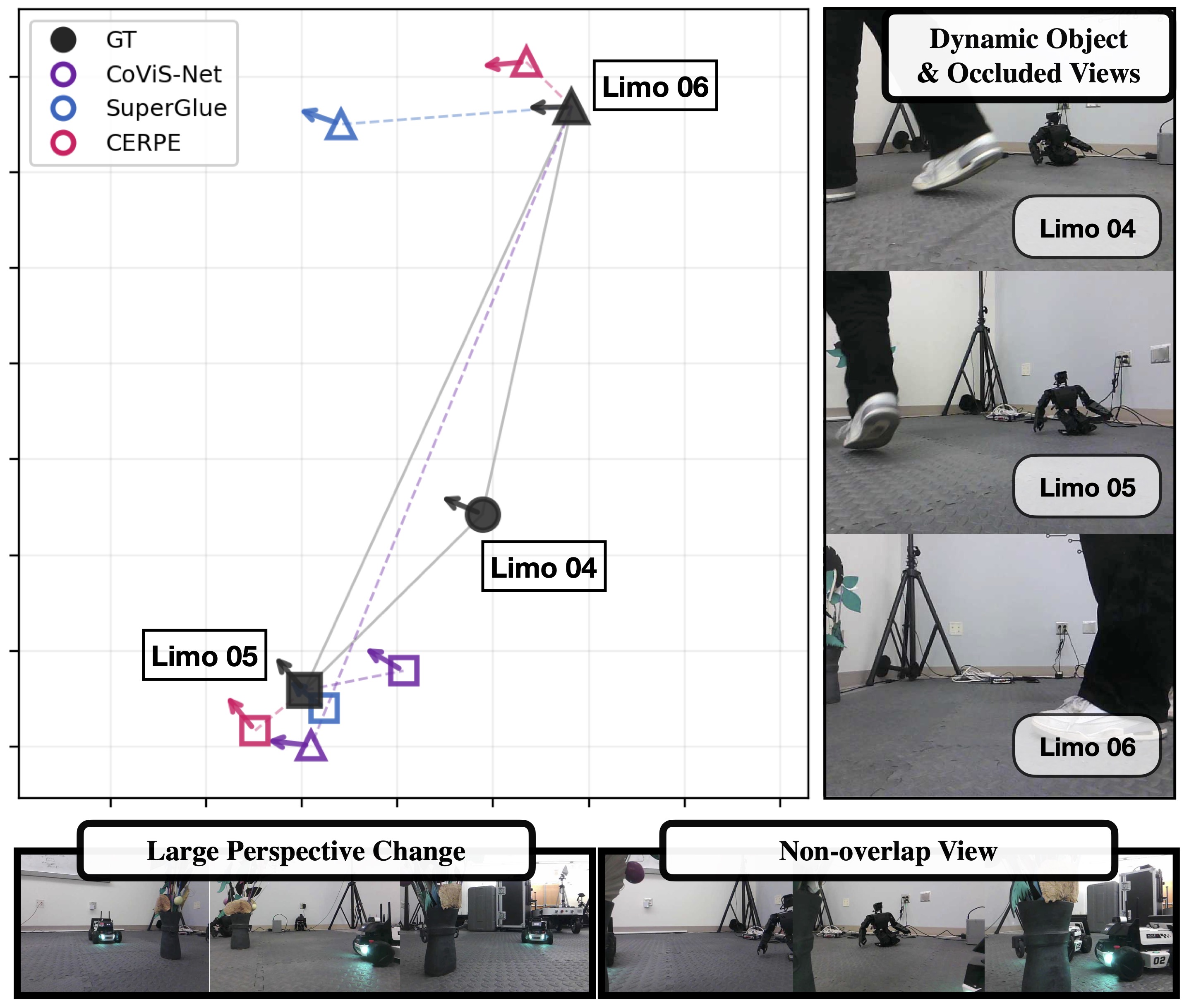}
    \caption{Qualitative results on physical robot teams in indoor scenarios with large perspective changes, non-overlapping views, and occlusions.}
    \label{fig:quali_robot}
    \vspace{-6pt}
\end{figure}

To evaluate physical-robot pose estimation beyond simulation and driving scenarios, we deploy \modelname on a team of three Agilex Limo Pro robots in indoor environments. Unlike the CAD scenarios where robots move along road-constrained trajectories, the indoor robot scenes feature close-range interactions with frequent transitions between overlapping and non-overlapping views, occluded observations caused by surrounding objects, and dynamic disturbances. As shown in Fig.~\ref{fig:quali_robot}, our CERPE predictions (red) are closer to the reference poses (black) for all three robots, whereas Co-VisNet (purple) and SuperGlue (blue) exhibit notable deviations, particularly for the distant Limo~06 where large perspective changes and dynamic occlusions are present.

As shown in Table~\ref{tab:main_results}, NOPE fails entirely on this dataset, as the indoor scenes lack the structured object categories (e.g., vehicles, pedestrians) that its object-level matching relies on. SuperGlue achieves the lowest translation error of 0.12\,m with ground-truth scale and succeeds on 376 out of 382 pairs at 98.4\%, benefiting from the close-range setting where sufficient feature correspondences are available. However, it still fails on 6 pairs and cannot recover metric scale. Co-VisNet produces estimates for all pairs but with substantially higher errors of 0.88\,m position and 0.257\,rad rotation, indicating limited generalization to the indoor domain.
Our \modelname achieves the lowest rotation error of 0.086\,rad, outperforming SuperGlue at 0.165\,rad by 48\% and Co-VisNet at 0.257\,rad by 67\%. In translation, \modelname w/o EM achieves the best predicted-scale result of 0.45\,m and ground-truth-scale result of 0.09\,m, while maintaining 100\% success. These results suggest that our approach transfers from driving scenarios to close-range indoor robot settings.

Finally, we run an online deployment test on 2--3 robots to check execution, communication, and controller integration; because motion-capture coverage does not span the full operating region, quantitative physical-robot accuracy is evaluated on the sampled mocap-covered scenarios above rather than from this closed-loop test. All computation runs on a laptop with an Intel i9-14900HX CPU and an Nvidia RTX 4080 GPU, and robots communicate over a shared wireless network using $640\times480$ JPEG-compressed images. The descriptor-only message is 16.9\,KB; overlap-triggered image exchange adds a compressed image averaging 26\,KB in our setup.
The full \modelname runs in 438\,ms for two robots and 780\,ms for three robots, while a 50\,Hz low-level controller consumes the latest pose estimate between perception updates.

\section{Conclusion}
We introduced CERPE, a communication-efficient framework for metric relative pose estimation under ephemeral encounters. CERPE coordinates existing vision foundation models through descriptor-gated image exchange, metric scale recovery, and ego-motion propagation from the latest direct pose anchor. Across CARLA, OpenMars, and indoor robot evaluations, CERPE maintains a 100\% pose-estimation success rate and improves accuracy over selected baselines in several settings without relying on GPS or maps. The results also show that propagation preserves continuity during missing-overlap intervals but can accumulate scale and translation drift when direct overlap constraints are strong. Direct inter-robot estimates therefore act as intermittent geometric anchors, while bounded-memory ego-motion bridges the intervals between them; descriptor screening remains separate from pose inference, so raw RGB transmission occurs only after gate activation.

Several limitations remain. Scale quality directly affects propagation and relative pose consistency, motivating additional metric cues such as stereo, LiDAR, or inertial sensing. The current shared-compute prototype targets small teams; larger teams require lighter or distributed inference. Finally, CERPE has no back-end optimization, and factor-graph refinement over keyframes and encounter anchors could reduce long-term drift.

\newpage
\bibliography{collabPerception}

@inproceedings{kendall2015posenet,
  title={Posenet: A convolutional network for real-time 6-dof camera relocalization},
  author={Kendall, Alex and Grimes, Matthew and Cipolla, Roberto},
  booktitle={Proceedings of the IEEE international conference on computer vision},
  pages={2938--2946},
  year={2015}
}

@inproceedings{zhang2022relpose,
  title={Relpose: Predicting probabilistic relative rotation for single objects in the wild},
  author={Zhang, Jason Y and Ramanan, Deva and Tulsiani, Shubham},
  booktitle={European Conference on Computer Vision},
  pages={592--611},
  year={2022},
  organization={Springer}
}

@inproceedings{lin2024relpose++,
  title={Relpose++: Recovering 6d poses from sparse-view observations},
  author={Lin, Amy and Zhang, Jason Y and Ramanan, Deva and Tulsiani, Shubham},
  booktitle={2024 International Conference on 3D Vision (3DV)},
  pages={106--115},
  year={2024},
  organization={IEEE}
}

@inproceedings{liu2020who2com,
  title={Who2com: Collaborative perception via learnable handshake communication},
  author={Liu, Yen-Cheng and Tian, Junjiao and Ma, Chih-Yao and Glaser, Nathan and Kuo, Chia-Wen and Kira, Zsolt},
  booktitle={2020 IEEE International Conference on Robotics and Automation (ICRA)},
  pages={6876--6883},
  year={2020},
  organization={IEEE}
}

@inproceedings{
nam2024diffmatch,
title={Diffusion Model for Dense Matching},
author={Jisu Nam and Gyuseong Lee and Sunwoo Kim and Hyeonsu Kim and Hyoungwon Cho and Seyeon Kim and Seungryong Kim},
booktitle={The Twelfth International Conference on Learning Representations},
year={2024},
url={https://openreview.net/forum?id=Zsfiqpft6K}
}

@inproceedings{wang2023posediffusion,
  title={Posediffusion: Solving pose estimation via diffusion-aided bundle adjustment},
  author={Wang, Jianyuan and Rupprecht, Christian and Novotny, David},
  booktitle={Proceedings of the IEEE/CVF International Conference on Computer Vision},
  pages={9773--9783},
  year={2023}
}

@inproceedings{sun2021loftr,
  title={LoFTR: Detector-free local feature matching with transformers},
  author={Sun, Jiaming and Shen, Zehong and Wang, Yuang and Bao, Hujun and Zhou, Xiaowei},
  booktitle={Proceedings of the IEEE/CVF conference on computer vision and pattern recognition},
  pages={8922--8931},
  year={2021}
}

@inproceedings{qin2022geometric,
  title={Geometric transformer for fast and robust point cloud registration},
  author={Qin, Zheng and Yu, Hao and Wang, Changjian and Guo, Yulan and Peng, Yuxing and Xu, Kai},
  booktitle={Proceedings of the IEEE/CVF conference on computer vision and pattern recognition},
  pages={11143--11152},
  year={2022}
}

@article{jiang2023se,
  title={Se (3) diffusion model-based point cloud registration for robust 6d object pose estimation},
  author={Jiang, Haobo and Salzmann, Mathieu and Dang, Zheng and Xie, Jin and Yang, Jian},
  journal={Advances in Neural Information Processing Systems},
  volume={36},
  pages={21285--21297},
  year={2023}
}

@article{miller2021any,
  title={Any way you look at it: Semantic crossview localization and mapping with lidar},
  author={Miller, Ian D and Cowley, Anthony and Konkimalla, Ravi and Shivakumar, Shreyas S and Nguyen, Ty and Smith, Trey and Taylor, Camillo Jose and Kumar, Vijay},
  journal={IEEE Robotics and Automation Letters},
  volume={6},
  number={2},
  pages={2397--2404},
  year={2021},
  publisher={IEEE}
}

@inproceedings{gao2020long,
  title={Long-term loop closure detection through visual-spatial information preserving multi-order graph matching},
  author={Gao, Peng and Zhang, Hao},
  booktitle={Proceedings of the AAAI Conference on Artificial Intelligence},
  volume={34},
  number={06},
  pages={10369--10376},
  year={2020}
}

@inproceedings{cui2022coopernaut,
  title={Coopernaut: End-to-End Driving with Cooperative Perception for Networked Vehicles},
  author={Cui, Jiaxun and Qiu, Hang and Chen, Dian and Stone, Peter and Zhu, Yuke},
  booktitle={IEEE/CVF Conference on Computer Vision and Pattern Recognition (CVPR)},
  year={2022}
}

@article{cuturi2013sinkhorn,
  title={Sinkhorn distances: Lightspeed computation of optimal transport},
  author={Cuturi, Marco},
  journal={Advances in neural information processing systems},
  volume={26},
  year={2013}
}

@article{fischler1981random,
  title={Random sample consensus: a paradigm for model fitting with applications to image analysis and automated cartography},
  author={Fischler, Martin A and Bolles, Robert C},
  journal={Communications of the ACM},
  volume={24},
  number={6},
  pages={381--395},
  year={1981},
  publisher={ACM New York, NY, USA}
}

@inproceedings{zhang2024diffglue,
  title={Diffglue: Diffusion-aided image feature matching},
  author={Zhang, Shihua and Ma, Jiayi},
  booktitle={Proceedings of the 32nd ACM International Conference on Multimedia},
  pages={8451--8460},
  year={2024}
}

@inproceedings{leroy2024grounding,
  title={Grounding image matching in 3d with mast3r},
  author={Leroy, Vincent and Cabon, Yohann and Revaud, J{\'e}r{\^o}me},
  booktitle={European conference on computer vision},
  pages={71--91},
  year={2024},
  organization={Springer}
}

@inproceedings{gao2023visual,
  title={Visual, spatial, geometric-preserved place recognition for cross-view and cross-modal collaborative perception},
  author={Gao, Peng and Liang, Jing and Shen, Yu and Son, Sanghyun and Lin, Ming C},
  booktitle={2023 IEEE/RSJ International Conference on Intelligent Robots and Systems (IROS)},
  pages={11079--11086},
  year={2023},
  organization={IEEE}
}

@article{lifelong_mapf, 
title={Lifelong Multi-Agent Path Finding in Large-Scale Warehouses}, 
journal={Proceedings of the AAAI Conference on Artificial Intelligence}, 
author={Li, Jiaoyang and Tinka, Andrew and Kiesel, Scott and Durham, Joseph W. and Kumar, T. K. Satish and Koenig, Sven} }

@inproceedings{xu2022v2xvit,
 author = {Runsheng Xu and Hao Xiang and Zhengzhong Tu and Xin Xia and Ming-Hsuan Yang and Jiaqi Ma},
 title = {V2X-ViT: Vehicle-to-Everything Cooperative Perception with Vision Transformer},
 booktitle={Proceedings of the European Conference on Computer Vision (ECCV)},
 year = {2022}}

@inproceedings{v2vnet,
  title={V2vnet: Vehicle-to-vehicle communication for joint perception and prediction},
  author={Wang, Tsun-Hsuan and Manivasagam, Sivabalan and Liang, Ming and Yang, Bin and Zeng, Wenyuan and Urtasun, Raquel},
  booktitle={European conference on computer vision},
  pages={605--621},
  year={2020},
  organization={Springer}
}

@inproceedings{gao2021regularized,
  title={Regularized graph matching for correspondence identification under uncertainty in collaborative perception},
  author={Gao, Peng and Guo, Rui and Lu, Hongsheng and Zhang, Hao Zhang},
  booktitle={Robotics science and systems},
  year={2021}
}

@inproceedings{wang2025vggt,
  title={VGGT: Visual Geometry Grounded Transformer},
  author={Wang, Jianyuan and Chen, Minghao and Karaev, Nikita and Vedaldi, Andrea and Rupprecht, Christian and Novotny, David},
  booktitle={Proceedings of the IEEE/CVF Conference on Computer Vision and Pattern Recognition},
  year={2025}
}

@article{gyagenda2022review,
  title={A review of GNSS-independent UAV navigation techniques},
  author={Gyagenda, Nasser and Hatilima, Jasper V and Roth, Hubert and Zhmud, Vadim},
  journal={Robotics and Autonomous Systems},
  volume={152},
  pages={104069},
  year={2022},
  publisher={Elsevier}
}

@article{tian2022kimera,
  title={Kimera-multi: Robust, distributed, dense metric-semantic slam for multi-robot systems},
  author={Tian, Yulun and Chang, Yun and Arias, Fernando Herrera and Nieto-Granda, Carlos and How, Jonathan P and Carlone, Luca},
  journal={IEEE Transactions on Robotics},
  volume={38},
  number={4},
  year={2022},
  publisher={IEEE}
}

@inproceedings{cieslewski2018data,
  title={Data-efficient decentralized visual SLAM},
  author={Cieslewski, Titus and Choudhary, Siddharth and Scaramuzza, Davide},
  booktitle={2018 IEEE international conference on robotics and automation (ICRA)},
  pages={2466--2473},
  year={2018},
  organization={IEEE}
}

@article{cao2024multi,
  title={Multi-robot object SLAM using distributed variational inference},
  author={Cao, Hanwen and Shreedharan, Sriram and Atanasov, Nikolay},
  journal={IEEE Robotics and Automation Letters},
  year={2024},
  publisher={IEEE}
}

@inproceedings{detone2018superpoint,
  title={Superpoint: Self-supervised interest point detection and description},
  author={DeTone, Daniel and Malisiewicz, Tomasz and Rabinovich, Andrew},
  booktitle={Proceedings of the IEEE conference on computer vision and pattern recognition workshops},
  pages={224--236},
  year={2018}
}

@inproceedings{sarlin2020superglue,
  title={Superglue: Learning feature matching with graph neural networks},
  author={Sarlin, Paul-Edouard and DeTone, Daniel and Malisiewicz, Tomasz and Rabinovich, Andrew},
  booktitle={Proceedings of the IEEE/CVF conference on computer vision and pattern recognition},
  pages={4938--4947},
  year={2020}
}

@inproceedings{zhan2024imatching,
  title={iMatching: Imperative correspondence learning},
  author={Zhan, Zitong and Gao, Dasong and Lin, Yun-Jou and Xia, Youjie and Wang, Chen},
  booktitle={European Conference on Computer Vision},
  pages={183--200},
  year={2024},
  organization={Springer}
}

@inproceedings{salad,
  title={Optimal transport aggregation for visual place recognition},
  author={Izquierdo, Sergio and Civera, Javier},
  booktitle={CVPR},
  year={2024}
}

@inproceedings{mars,
  title={Multiagent multitraversal multimodal self-driving: Open mars dataset},
  author={Li, Yiming and Li, Zhiheng and Chen, Nuo and Gong, Moonjun and Lyu, Zonglin and Wang, Zehong and Jiang, Peili and Feng, Chen},
  booktitle={Proceedings of the IEEE/CVF Conference on Computer Vision and Pattern Recognition},
  pages={22041--22051},
  year={2024}
}

@inproceedings{roma,
	title={Deep Regression on Manifolds: a {3D} Rotation Case Study},
	author={Br{\'e}gier, Romain},
	journal={2021 International Conference on 3D Vision (3DV)},
	year={2021}
}

@inproceedings{wang2024dust3r,
  title={Dust3r: Geometric 3d vision made easy},
  author={Wang, Shuzhe and Leroy, Vincent and Cabon, Yohann and Chidlovskii, Boris and Revaud, Jerome},
  booktitle={Proceedings of the IEEE/CVF Conference on Computer Vision and Pattern Recognition},
  pages={20697--20709},
  year={2024},
}

@misc{nope2025,
      title={Non-Overlap-Aware Egocentric Pose Estimation for Collaborative Perception in Connected Autonomy}, 
      author={Hong Huang and Dongkuan Xu and Hao Zhang and Peng Gao},
      year={2025},
      eprint={2506.14180},
      archivePrefix={arXiv},
      primaryClass={cs.RO},
}

@misc{covis2024,
      title={CoViS-Net: A Cooperative Visual Spatial Foundation Model for Multi-Robot Applications}, 
      author={Jan Blumenkamp and Steven Morad and Jennifer Gielis and Amanda Prorok},
      year={2024},
      eprint={2405.01107},
      archivePrefix={arXiv},
      primaryClass={cs.RO},
}

@inproceedings{
pi3,
title={\${\textbackslash}pi{\textasciicircum}3\$: Permutation-Equivariant Visual Geometry Learning},
author={Yifan Wang and Jianjun Zhou and Haoyi Zhu and Wenzheng Chang and Yang Zhou and Zizun Li and Junyi Chen and Jiangmiao Pang and Chunhua Shen and Tong He},
booktitle={The Fourteenth International Conference on Learning Representations},
year={2026},
}

@misc{mapany,
      title={MapAnything: Universal Feed-Forward Metric 3D Reconstruction}, 
      author={Nikhil Keetha and Norman Müller and Johannes Schönberger and Lorenzo Porzi and Yuchen Zhang and Tobias Fischer and Arno Knapitsch and Duncan Zauss and Ethan Weber and Nelson Antunes and Jonathon Luiten and Manuel Lopez-Antequera and Samuel Rota Bulò and Christian Richardt and Deva Ramanan and Sebastian Scherer and Peter Kontschieder},
      year={2025},
      eprint={2509.13414},
      archivePrefix={arXiv},
      primaryClass={cs.CV},
      url={https://arxiv.org/abs/2509.13414}, 
}

@article{hu2024metric3dv2,
  title={Metric3d v2: A versatile monocular geometric foundation model for zero-shot metric depth and surface normal estimation},
  author={Hu, Mu and Yin, Wei and Zhang, Chi and Cai, Zhipeng and Long, Xiaoxiao and Chen, Hao and Wang, Kaixuan and Yu, Gang and Shen, Chunhua and Shen, Shaojie},
  journal={IEEE Transactions on Pattern Analysis and Machine Intelligence},
  year={2024},
  publisher={IEEE}
}

@inproceedings{carla,
  title = {{CARLA}: {An} Open Urban Driving Simulator},
  author = {Alexey Dosovitskiy and German Ros and Felipe Codevilla and Antonio Lopez and Vladlen Koltun},
  booktitle = {Proceedings of the 1st Annual Conference on Robot Learning},
  pages = {1--16},
  year = {2017}
}

@article{gao2024survey,
  title={A survey of collaborative perception in intelligent vehicles at intersections},
  author={Gao, Xin and Zhang, Xinyu and Lu, Yiguo and Huang, Yuning and Yang, Lei and Xiong, Yijin and Liu, Peng},
  journal={IEEE Transactions on Intelligent Vehicles},
  year={2024},
  publisher={IEEE}
}
\bibliographystyle{IEEEtran}
\end{document}